# Direct White Matter Bundle Segmentation using Stacked U-Nets

Jakob Wasserthal, Peter F. Neher, Fabian Isensee, and Klaus H. Maier-Hein

Medical Image Computing Group,
German Cancer Research Center (DKFZ), Heidelberg, Germany
`j.wasserthal@dkfz.de`

**Abstract.** The state-of-the-art method for automatically segmenting white matter bundles in diffusion-weighted MRI is tractography in conjunction with streamline cluster selection. This process involves long chains of processing steps which are not only computationally expensive but also complex to setup and tedious with respect to quality control. Direct bundle segmentation methods treat the task as a traditional image segmentation problem. While they so far did not deliver competitive results, they can potentially mitigate many of the mentioned issues. We present a novel supervised approach for direct tract segmentation that shows major performance gains. It builds upon a stacked U-Net architecture which is trained on manual bundle segmentations from Human Connectome Project subjects. We evaluate our approach *in vivo* as well as *in silico* using the ISMRM 2015 Tractography Challenge phantom dataset. We achieve human segmentation performance and a major performance gain over previous pipelines. We show how the learned spatial priors efficiently guide the segmentation even at lower image qualities with little quality loss.

## 1 Introduction

Diffusion MRI is an important tool in the study of the brain's white matter with tractography being the state of the art method for reconstructing white matter pathways. Tractography typically results in thousands of streamlines that require filtering for false positive removal and generation of anatomically meaningful bundle segmentations [14, 12]. In this context, streamline selection can be performed using different approaches. Interactively drawn regions of interest (ROIs) are very commonly used but are quite time consuming and require experts. Small changes alter the resulting bundles significantly and limit reproducibility across subjects and human experts [21, 18]. Automation can be achieved using atlas-guided approaches [17, 7] or selection schemes based on gray matter parcellations [22]. Atlas information can as well be directly integrated into the tractography process in form of prior information [23, 10, 2]. Such automated approaches involve quite complex registration and/or segmentation procedures, i.e. registration of T1-weighted and diffusion-weighted images, registration across individuals and segmentation of T1-weighted images. In combination with the tractography itself they add up to long chains of processing steps that are difficult to oversee and configure and that involve computing-intensive operations in streamline space.

*Direct bundle segmentation* methods potentially solve many of these issues by circumventing tractography and treating the task as a traditional image segmentation problem. Previous studies applied level sets, active contours or related markov random field-based methods to this problem [11, 5, 8, 9, 4]. These methods require the specification of initial ROIs corresponding to specific tracts of interest. Prior information about the tracts can be incorporated using atlas information [1, 13] or deformable tract templates [3].

*Supervised learning* has shown great potential in the derivation of spatial priors from training data, while avoiding the issues that atlas based methods encompass. However, this potential has been barely explored yet when it comes to direct bundle segmentation. Ratnarajah et al. published, to our knowledge, the only study towards this direction, using k-NN based classification in Riemannian diffusion-tensor spaces to label white matter fiber bundles in neonatal brain images [15].

We propose a novel approach for direct white matter bundle segmentation using supervised learning on basis of a stacked U-Net architecture that receives the peaks of the fiber orientation distribution functions (fODF) as input. This results in a very detailed 3D model for each bundle and does not rely on long chains of processing steps involving atlas based priors or fiber tractography. To train our model we interactively [18] labeled 30 subjects from the Human Connectome Project (HCP), yielding a high quality dataset of the subject specific morphology of these bundles. We evaluate our approach qualitatively and quantitatively on high and low quality *in vivo* data and show that it is capable of segmenting larger and even very thin bundles with high quality in a couple of seconds. Moreover, we compare our approach to the results of the ISMRM 2015 Tractography Challenge [12] and to segmentation results previously reported in the literature. On *in vivo* data our method reaches human performance, thus providing an efficient and precise solution for automatic segmentation of white matter fiber bundles.

## 2 Methods

**Model** Our convolutional neural network is based on the U-Net architecture [16]. Because of the high resolution of our data it would not be memory efficient to use the entire 3D image as input. Thus we propose using a set of 2D slices as input. We train three individual networks, one for each spatial axis x, y and z. Training slices for the three networks are thus sampled from the x-y, y-z, and z-x plane, respectively. This results in three predictions per voxel. We concatenate these predictions to create a new image with three channels, on which we train a fourth U-Net that yields the final prediction.

We considered several different types of input features for our network. Since using the raw image values or parametric representations of the signal, such as spherical harmonics coefficients, would result in a very large number of features and a successively very high memory demand, we decided to use the voxel-wise principal fiber direction vector-elements as features. This reduces the number of features while preserving important information about the local white matter structure, e.g. about crossing fibers. To obtain the principal fiber directions, we used the constrained spherical deconvolution and peak extraction available in MRtrix [6, 19] with a maximum number of three peaks per voxel. This results in a total number of nine features per voxel. The feature images are processed slice by slice, where each slice has nine channels corresponding to the nine features. A binary segmentation of the bundle of interest serves as the training target. The output of our network is the probability for each voxel to belong to a certain bundle.

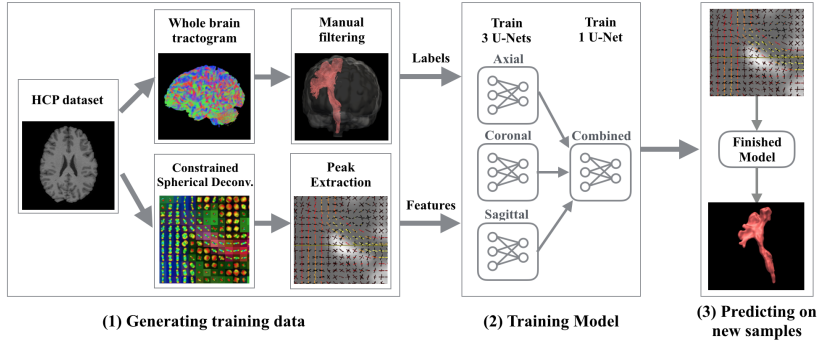

**Fig. 1.** Overview of our pipeline: (1) Generation of training data by manual filtering of tractograms and generation of features, (2) Training stacked U-Net, (3) Final model for segmentation of bundles

**Training** The following hyperparameters were used to train the network: a learning rate of 0.002, a batch size of 8, 70 training epochs and a dropout probability of 0.4. The learning rate is decreased by 3% per epoch. As loss function we chose the categorical crossentropy. Because of the great class imbalance (the fiber bundle makes up only a small part of the image) we weight the loss of bundle voxels higher than the loss of non-bundle voxels. The weighting corresponds to the inverse class frequency. This guides the learning process to focus on the fewer bundle voxels. This weighting is linearly decreased over the epochs. The input images were normalized to zero mean and a standard deviation of one. All hyperparameters were optimized on an independent validation datasets. The network weights of the epoch with the highest Dice score were used for testing.

**Datasets** Based on the results of the ISMRM tractography challenge, the organizers assigned each of the 25 tracts used during the challenge to one of three difficulty groups ("medium", "hard" and "very hard") [12]. To evaluate our approach we trained it separately for one tract out of each category: left superior longitudinal fasciculus (SLF) ("medium"), the right corticospinal tract (CST) ("hard") and anterior commissure (CA) ("very hard").

*HCP_highRes:* In our *in-vivo* experiments we used 30 subjects of the Human Connectome Project. The HCP diffusion weighted images were acquired with 1.25mm isotropic resolution and 270 gradient directions with three b-values ($b = 1000s/mm^2$, $b = 2000s/mm^2$, $b = 3000s/mm^2$). The reference segmentations of the three tracts needed for training and testing of our approach were created manually using the following pipeline: (1) performing standard whole brain tractography of all subjects using MRtrix [20], (2) manual extraction of the desired tracts using ROIs following a white matter atlas [18], (3) conversion of the resulting fiber bundles into binary segmentation images. Fiber tractography was performed using MRtrix multi-tissue constrained spherical deconvolution (CSD) of multi-shell data [6] and anatomically-constrained probabilistic streamline tractography [20], yielding 1 million fibers with a minimum fiber length of 80 mm per subject. The other parameters were kept at their default values. From the fODFs we extract three peaks as input features for our model using MRtrix *sh2peaks*.

*HCP_lowRes:* To test our method in a clinically more plausible setting and to be comparable to the ISMRM Tractography Challenge dataset, we sampled the high resolution dataset to 2mm isotropic resolution and removed all $b = 2000s/mm^2$ and $b = 3000s/mm^2$ gradient directions. From the remaining gradient directions we sampled 32 gradients evenly distributed over the entire sphere. For estimating the fODFs we use MRtrix constrained spherical deconvolution [19] and also extract three peaks as input features.

*ISMRM Challenge:* We also test our model on the ISMRM Tractography Challenge phantom. The phantom was denoised and corrected for distortions using MRtrix. fODFs and peaks were extracted the same way as for HCP_lowRes.

**Experiments** We performed three experiments: (1) Training and testing on HCP_highRes, (2) Training and testing on HCP_lowRes and (3) Training on HCP_lowRes and testing on the ISMRM Tractography Challenge phantom.
We split our dataset into 20 train subjects, 5 validation subjects for hyperparameter optimization and 5 test subjects for final evaluation. As metric we use the Dice score. To compare our model to human performance a second expert rater segmented the test subjects. The inter-rater variations of the two experts are called "Human" performance in section 3. We show results for our stacked U-Net architecture and compare to a plain U-Net that is only trained on 2D slices from one plain.

## 3  Results

**In vivo** On the HCP high resolution dataset (HCP_highRes) our method achieves human performance on all bundles. On the CST und SLF it even outperforms the human results by a small margin (Figure 2). Our stacked U-Net architecture shows a clear performance improvement compared to a plain U-Net, especially on the low resolution data. On low resolution data the performance of our model decreases only slightly. This makes it very promising for clinical settings, where no high resolution data is available. The qualitative evaluation confirms that our model manages very well to segment the ground truth shapes (Figure 3).

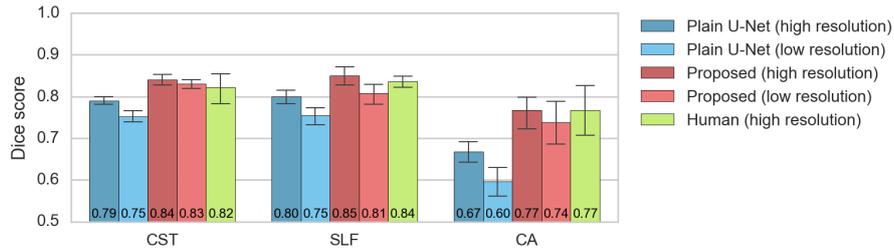

**Fig. 2.** Results of a plain U-Net and our proposed stacked U-Net on the HCP high and low resolution datasets in comparison to human performance (second rater).

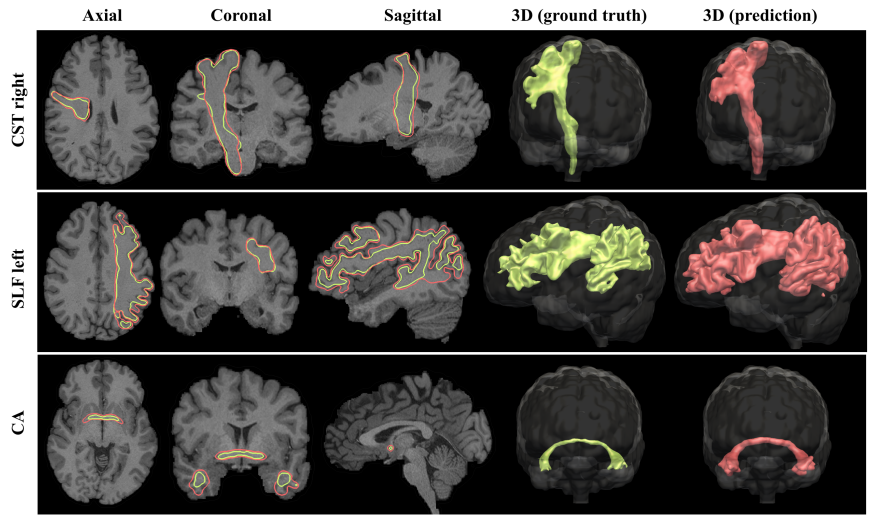

**Fig. 3.** Ground truth (green) and prediction of our model (red) on a randomly chosen subject from the test set of the HCP high resolution dataset.

**Phantom** To evaluate our model against the phantom of the ISMRM Tractography Challenge we trained the model on the HCP_lowRes dataset and tested it on the phantom. We also tested it on a version of the phantom that contains no simulated artifacts and noise ("no artifacts"). Figure 4 shows the results.

On the CST our model is part of the top three submission and is able to find the lateral projections of the CST, which other methods often miss. On the SLF our model shows competitive results. On the CA our model is not able to properly reconstruct the entire bundle. However, it finds more than most of the challenge submissions. As expected on the "no artifacts" phantom we achieve slightly better results.

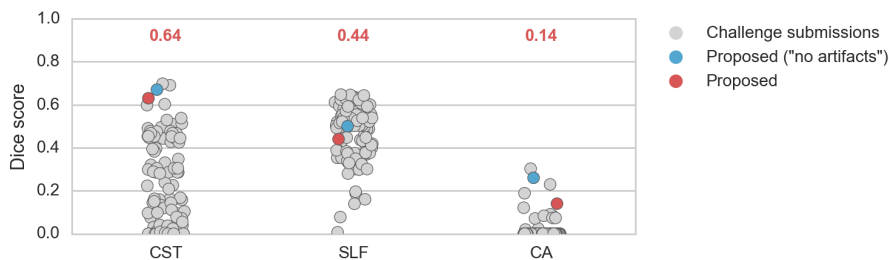

**Fig. 4.** Results on the ISMRM 2015 Tractography Challenge phantom. *Proposed ("no artifacts")*: Our model on phantom without artifacts. *Proposed*: Our model on original phantom. *Challenge submissions*: Submissions of all the challenge participants. For better visualization we use a small random displacement in x direction in the plot.

## 4 Discussion

We presented a new stacked U-Net architecture for direct segmentation of anatomically meaningful white matter bundles. With this direct method we avoid errors being accumulated by using long processing chains involving atlas registrations and fiber tractography, where each step is an additional source of inaccuracy. Furthermore, this enables us to process a whole brain in less than five seconds using a recent GPU, which is very fast compared to many other commonly used methods for fiber bundle segmentation or tractography.

*In vivo* we show that our method achieves near-human or even better-than-human performance and that the segmentation quality decreases only slightly even when using data of much lower quality. Furthermore, we show that our approach of stacked U-Nets with different spatial specializations improves the overall segmentation results significantly compared to a single U-Net.

A comparison to segmentation results previously reported in the literature is difficult due to the diversity of employed datasets, analyzed anatomical structures and employed evaluation metrics. In general it is worth noting that other works focused on quite large and prominent bundles, such as the arcuate fasciculus, corticospinal tract (CST), inferior fronto-occipital fasciculus or the fornix [10,15], while our approach also yields very good results for thin anatomical structures such as the anterior commissure. When comparing our Dice scores for singular anatomical structures we can see that our method outperforms previously reported results: right CST (0.84 [proposed] vs < 0.8 [1], ∼ 0.72 [15], < 0.65 [10], < 0.51 [10]) and left SLF (0.85 [proposed] vs < 0.8 [1]). Kasenburg et al. [10] only report true positive scores so we assumed zero false negatives to obtain an upper bound for the respective Dice scores.

On the ISMRM 2015 Tractography Challenge phantom our network achieves competitive results to other challenge submissions. However, performance on the phantom is clearly lower as compared to the *in vivo* experiments. This is also true for the phantom without noise and artifacts, which suggests that the better performance on the HCP data is not due to better signal quality. We assume that this is caused by structural differences between the HCP data and the phantom: Bundles are defined smaller in the phantom as compared to the *in vivo* reference segmentations and the phantom does not contain many structures that are present *in vivo*, which might be necessary for our approach to delimit the target bundles. Training on *in silico* data with a reference similar to the ISMRM phantom will probably alleviate this issue.

Our future efforts will focus on the inclusion of additional fiber bundles, such as the bundles used for the ISMRM 2015 tractography challenge. Furthermore, we are currently working on methodological extensions to train a single stacked U-Net jointly on multiple bundles as well as on the closer assessment of alternatively preprocessed or non-processed input data. The goal is to provide an easy-to-use framework for public use to automatically segment a comprehensive set of well known fiber tracts as well as the provision of pretrained models for processing the datasets of the Human Connectome Project. Applications include tasks such as tractometry as well as a posteriori filtering of tractograms or the provision of subject specific a priori knowledge for tractography.